%% file: main.tex
\documentclass[10pt,twocolumn,letterpaper]{article}

\usepackage{cvpr}              
\usepackage[accsupp]{axessibility} 
\usepackage{amssymb}
\usepackage{pifont}
\newcommand{\xmark}{\ding{55}}%
\input{preamble}

\definecolor{cvprblue}{rgb}{0.21,0.49,0.74}
\usepackage[pagebackref,breaklinks,colorlinks,citecolor=cvprblue]{hyperref}

\def\paperID{16045} 
\def\confName{CVPR}
\def\confYear{2024}

\title{SelfPose3d: Self-Supervised Multi-Person Multi-View 3d Pose Estimation}

\author{Vinkle Srivastav$^{1,2*}$ \quad Keqi Chen$^{1*}$ \quad  Nicolas Padoy$^{1,2}$  \vspace{0.3em} \\
{\normalsize $^1$University of Strasbourg, CNRS, INSERM, ICube, UMR7357, Strasbourg, France} \\
{\normalsize $^2$IHU Strasbourg, France} \\
{\tt\small srivastav@unistra.fr \quad  keqi.chen@unistra.fr  \quad npadoy@unistra.fr}
}

\begin{document}
\maketitle
\def\thefootnote{*}\footnotetext{co-first authors with equal contributions.}\def\thefootnote{\arabic{footnote}}
\input{sec/0_abstract}    
\input{sec/1_intro}
\input{sec/2_related}
\input{sec/3_method}

\input{sec/4_experiments}
\input{sec/5_conclusion}
\section{Acknowledgements}
This work was partially supported by French state funds managed by the ANR under references ANR-20-CHIA-0029-01 (National AI Chair AI4ORSafety), ANR-10-IAHU-02 (IHU Strasbourg), ANR-18-CE45-0011-03 (OptimiX), and by BPI France (project 5G-OR). This work was also granted access to the servers/HPC resources managed by CAMMA, IHU Strasbourg, Unistra Mesocentre, and GENCI-IDRIS [Grant 2021-AD011011638R3].

{
    \small
    \bibliographystyle{ieeenat_fullname}
    \bibliography{main}
}
\input{sec/X_suppl}


\end{document}

%% file: preamble.tex
\usepackage[dvipsnames]{xcolor}

\usepackage{bbm}
\usepackage{multirow}
\usepackage{makecell}

%% file: sec/0_abstract.tex
\begin{abstract}
We present a new self-supervised approach, \emph{SelfPose3d}, for estimating 3d poses of multiple persons from multiple camera views. Unlike current state-of-the-art fully-supervised methods, our approach does not require any 2d or 3d ground-truth poses and uses only the multi-view input images from a calibrated camera setup and 2d pseudo poses generated from an \emph{off-the-shelf} 2d human pose estimator. We propose two self-supervised learning objectives: self-supervised person localization in 3d space and self-supervised 3d pose estimation. We achieve self-supervised 3d person localization by training the model on synthetically generated 3d points, serving as 3d person root positions, and on the projected root-heatmaps in all the views. We then model the 3d poses of all the localized persons with a bottleneck representation, map them onto all views obtaining 2d joints, and render them using 2d Gaussian heatmaps in an end-to-end differentiable manner. Afterwards, we use the corresponding 2d joints and heatmaps from the pseudo 2d poses for learning. To alleviate the intrinsic inaccuracy of the pseudo labels, we propose an adaptive supervision attention mechanism to guide the self-supervision. Our experiments and analysis on three public benchmark datasets, including Panoptic, Shelf, and Campus, show the effectiveness of our approach, which is comparable to fully-supervised methods. Code is available at \url{https://github.com/CAMMA-public/SelfPose3D}.
\end{abstract}

%% file: sec/1_intro.tex
\section{Introduction}
\label{sec:intro}

\begin{figure*}[t!]
	\centering
	\begin{subfigure}[t]{0.43\textwidth}
		\centering
		\includegraphics[width=0.95\linewidth]{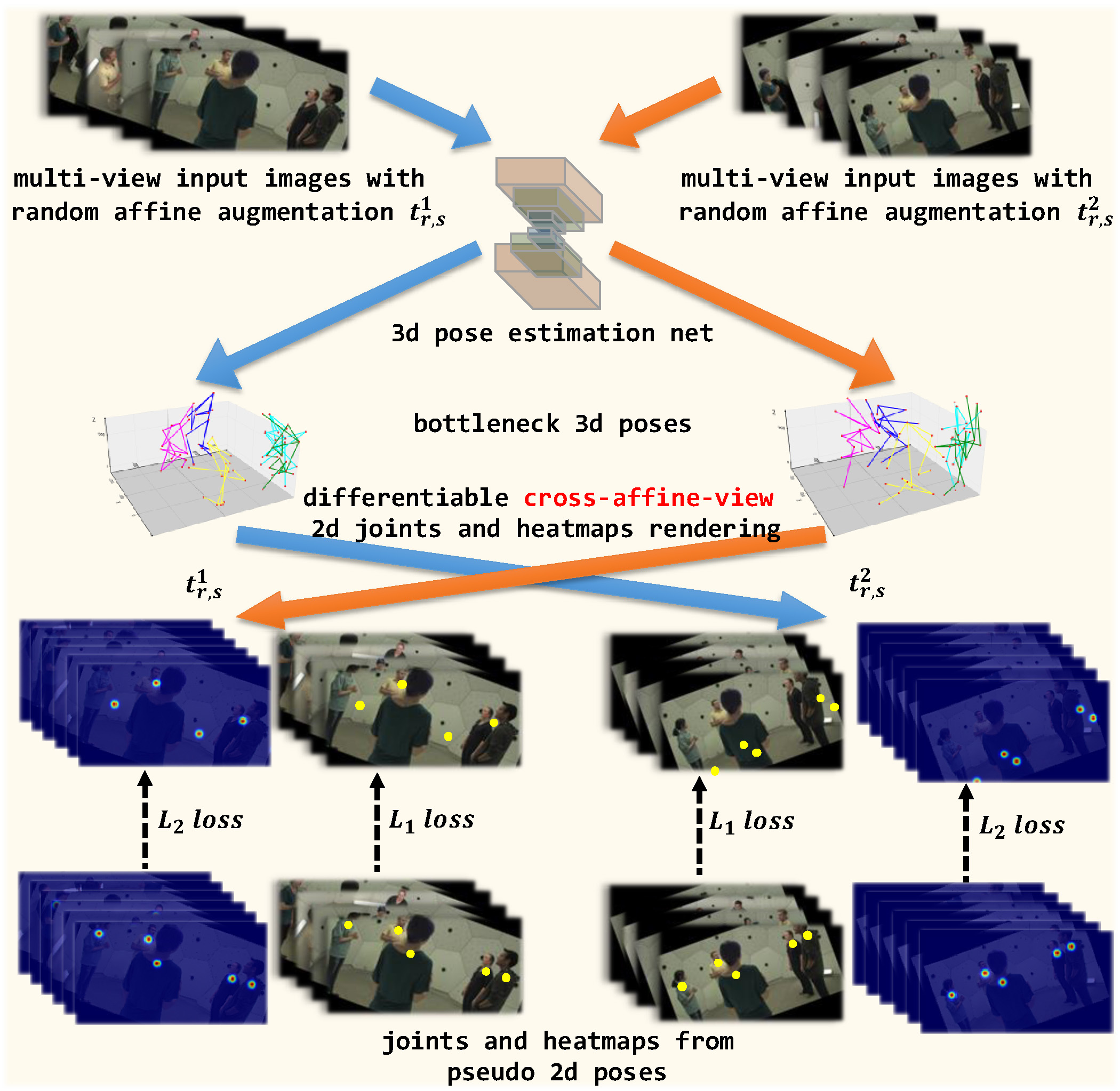}
	\end{subfigure}
	\begin{subfigure}[t]{0.56\textwidth}
		\centering
		\includegraphics[width=0.94\linewidth]{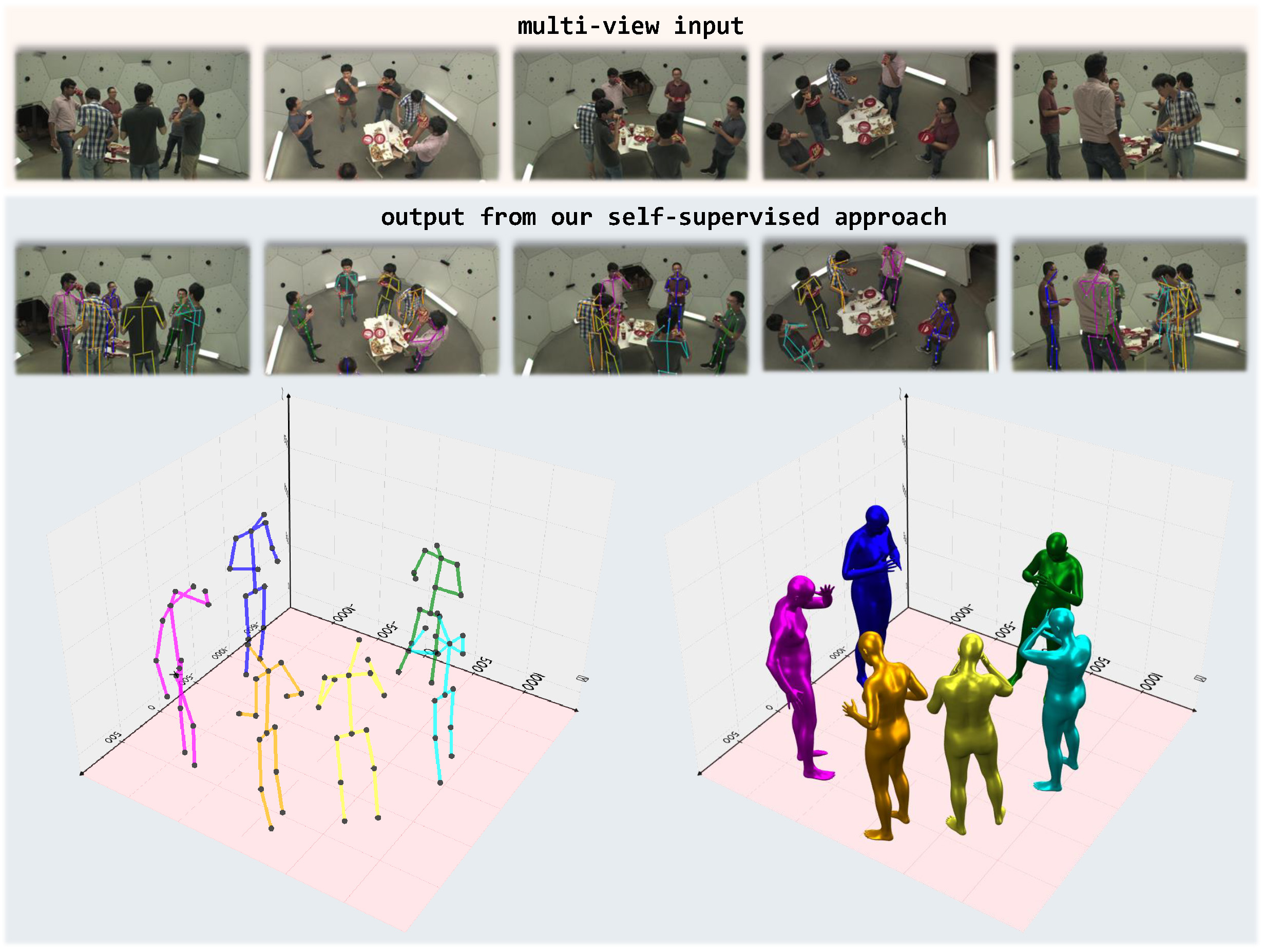}
	\end{subfigure}
	\caption{\small{Our self-supervised approach, called SelfPose3d, estimates multi-person 3d poses from multi-view images and pseudo 2d poses generated using an \emph{off-the-shelf} 2d human pose estimator. We propose a self-supervised learning objective that generates differentiable and geometrically constrained 2d joints and heatmaps across multiple views from bottleneck 3d poses. On the right, we show 3d pose outputs from our approach along with estimated body meshes (using SMPL body mesh fitting on 3d poses \cite{loper2015smpl,bogo2016keep}) and the projected 2d poses.}}
	\label{fig:intro-fig}
	\vspace{-3mm}
\end{figure*}

The task of estimating 3d poses for multiple persons using a few calibrated cameras is a challenging computer vision problem~\cite{tu2020voxelpose,reddy2021tessetrack,joo2015panoptic,dong2019fast,chen2020multi}. A significant part of this challenge lies in identifying and matching the same person across different camera views. The solutions developed so far generally use one of the two paradigms: \emph{learning-based methods} and \emph{optimization-based methods}. The learning-based methods develop novel deep-learning models and use 3d ground-truth poses for both training the models and establishing person correspondences across different views \cite{tu2020voxelpose,lin2021multi,zhang2021direct,wu2021graph,reddy2021tessetrack}. The accurate 3d ground-truth poses are typically generated using a dense camera system \cite{joo2015panoptic}. In contrast, the optimization-based methods formulate the 3d pose reconstruction as a mathematical optimization task, primarily focusing on aligning and matching the 2d poses across different camera views to infer 3d poses using triangulation within the framework of multi-view geometry \cite{joo2014map,joo2015panoptic,pirinen2019domes,dong2019fast,chen2020multi,kadkhodamohammadi2021generalizable}. The 2d poses are estimated using off-the-shelf 2d human pose detectors \cite{sun2019deep,cao2017realtime}. These methods apply geometric and spatial constraints in the optimization loop to ensure the anatomical plausibility and consistency of the inferred 3d poses. Although these methods do not require 3d ground-truth poses, their effectiveness is somewhat limited compared to the fully-supervised learning-based methods, see Table~\ref{tab:panoptic_main}.

In this paper, we explore the possibility of combining the strengths of both paradigms. Specifically, we investigate whether it's feasible to utilize a learning-based model for multi-view, multi-person 3D pose estimation and simultaneously eliminate its dependence on 3D ground-truth poses by incorporating geometric and appearance constraints, drawing inspiration from optimization-based methods.

We propose, \emph{SelfPose3d}, a self-supervised learning-based approach to estimate the 3d poses of multiple persons from a few calibrated cameras without using any 2d or 3d ground-truth poses. Our approach requires only \emph{2d pseudo poses} obtained using an off-the-shelf 2d pose detector \cite{sun2019deep}. Learning 3d poses without 3D ground-truth poses would require suitable supervisory signals to train a learning-based model. We follow the \emph{learning-by-projection} paradigm, where the main idea is to learn the 3d output by comparing the projected bottleneck 3d output against the 2d input features~\cite{chen2019learning}. 

We consider VoxelPose \cite{tu2020voxelpose} as a learning-based method and use its output 3d poses as a bottleneck representation. To recover the accurate underlying 3d poses, we propose using \emph{differentiable multi-view 2d representations} and \emph{cross-affine-view consistency}. In particular, given a multi-view input image, we apply two random affine augmentations and pass them to the VoxelPose. It generates the bottleneck 3d poses corresponding to each affine augmented multi-view image. To enforce the model to learn and reason in the spatial dimension, we project the bottleneck 3d poses onto each view, obtaining 2d joints, and rendering them into spatial 2d heatmap representations in an end-to-end differentiable way. We further put tight geometric constraints by cross-affine-view operation, \ie the bottleneck 3d poses from the 1st affine augmented multi-view image is mapped and rendered in the 2nd affine augmented multi-view image space and vice versa. Finally, we use the affine transformed 2d joints and heatmaps from the \emph{2d pseudo poses} to enable the geometrically constrained learning, with $L_1$ and $L_2$ losses respectively.

As the \emph{2d pseudo poses} contain non-negligible noises (mostly due to occlusions, see \Cref{fig:abl_1}), we propose \emph{adaptive supervision attention} to guide our model to focus on more reliable regions. We apply two strategies towards $L_1$ joint loss and $L_2$ heatmap loss; for $L_1$ joint loss supervision, we employ hard attention, where we ignore the one view with the largest absolute error for each multi-view image set; for $L_2$ heatmap loss supervision, we employ soft attention using a lighter backbone to process each view, obtaining same-size attention heatmaps. During $L_2$ loss computation, we compute the element-wise product of the attention heatmaps and the square error before averaging. To avoid obtaining zero attention, which the model tends to do, we add a regularization term, where we create tensors of all ones as the attention heatmap labels and use $L_2$ loss as the attention loss.

Finally, specific to our choice of learning-based method, \ie, VoxelPose, which uses a voxel-based 3d root localization model to localize the persons in space using ground-truth 3d root joints (mid-hip joint), we use a simple but effective strategy to localize persons in space. Specifically, we randomly place 3d points in 3d world-space and project them to each view using the given camera parameters, subsequently rendering the projected 2d points as heatmap representations. This generates a synthetic dataset containing 3d points (roots) and their corresponding rendered multi-view root-heatmaps. We then use this dataset to train a 3d root localization model that takes multi-view root-heatmaps as input and predicts the 3d roots as output. We further regularize the model by enforcing invariant constraints between pairs of affine augmented root-heatmaps coming from the real multi-view input.  

Evaluation on three 3d pose benchmarks datasets, Panoptic \cite{joo2015panoptic}, Shelf \cite{belagiannis20143d} and Campus \cite{belagiannis20143d}, along with extensive ablation studies on the Panoptic \cite{joo2015panoptic} dataset, show the effectiveness of our approach. Our approach reaches a performance comparable to learning-based fully-supervised approaches and performs significantly better than optimization-based approaches. Moreover, SMPL body mesh fitting \cite{loper2015smpl,bogo2016keep} on our estimated 3d poses generates geometrically plausible body shapes (see \Cref{fig:intro-fig} and \Cref{fig:qual_1}).

We summarize our contributions as follows: 1) We address the challenging multi-person multi-view 3d person pose estimation problem using a self-supervised approach without any 2d or 3d ground truth. 2) We propose \emph{self-supervised 3d pose estimation} by using a new method to recover geometrically constrained 2d joints and heatmap representations from the bottleneck 3d poses. 3) We propose \emph{adaptive supervision attention} to address the misinformation caused by the inaccurate pseudo labels. 4) We propose \emph{self-supervised 3d root localization} to estimate the 3d root location utilizing synthetic 3d roots and the corresponding rendered multi-view root heatmaps.

%% file: sec/2_related.tex
\section{Related work}
\label{sec:related}

In this section, we briefly review current works related to fully-supervised learning-based methods for 3d pose estimation, optimization-based methods for 3D pose estimation, and self-supervised learning.

\noindent\textbf{Fully-supervised methods:} Monocular 3d pose estimation \cite{zhou2017towards,Martinez17,mehta2017vnect,popa2017deep,sun2018integral,nie2019single,zhang2020inference,gong2021poseaug} is an ill-posed problem due to depth ambiguities as multiple 3d poses can produce same 2d pose projection. Having access to multi-view cameras can alleviate such depth ambiguities achieving the state-of-art results on benchmark datasets \cite{qiu2019cross,he2020epipolar,iskakov2019learnable,tu2020voxelpose,lin2021multi,zhang2021direct,wu2021graph,reddy2021tessetrack,gordon2021flex}. For single-person scenes, these approaches exploit multi-view geometry \cite{hartley2003multiple} to either fuse the visual features \cite{qiu2019cross, he2020epipolar}, perform triangulation on heatmaps \cite{iskakov2019learnable,remelli2020lightweight}, or use pictorial structural models for 3d reconstruction \cite{PavlakosZDD17,qiu2019cross}. The multi-person scene offers extra complexity due to the variability in the number of person in each view and the unknown cross-view correspondence. Existing multi-person multi-view approaches are based on volumetric paradigm \cite{tu2020voxelpose,reddy2021tessetrack,zhang2021voxeltrack}, or direct regression \cite{zhang2021direct} based on transformers \cite{vaswani2017attention,zhu2020deformable,carion2020end}. Despite their good performance, these approaches rely on ground-truth 3d poses, which are generated using dense camera systems \cite{joo2015panoptic}. 
%

\noindent\textbf{Optimization-based 3d pose estimation:} For the multi-person and multi-view scenario, optimization-based approaches use an \emph{off-the-shelf} person-id detector across all the views to solve the correspondence and triangulation problem and temporal refinement along with training a reinforcement learning agent to find the best camera locations for 3d pose reconstruction \cite{pirinen2019domes}. More recent approaches utilize multi-view 3d reconstruction in the optimization loop inferring 3D poses that are geometrically and spatially coherent \cite{dong2019fast,chen2020multi,kadkhodamohammadi2021generalizable}.

\noindent\textbf{{Self-supervised learning:}} Self-supervised learning can be broadly classified into self-supervised \emph{representation} learning and self-supervised \emph{task} learning. Self-supervised \emph{representation} learning aims to use large-scale unlabeled data to learn generic feature representations. The recent promising results from these approaches have started to surpass the fully-supervised baselines for various downstream tasks \cite{chen2020simple,he2020momentum,caron2021emerging,richemond2020byol}. Self-supervised task learning aims to learn a particular downstream task without using ground truth labels and has been applied to 2d pose estimation \cite{jakab2018unsupervised,jakab2020self}, single-person 3d pose estimation \cite{kocabas2019self,kundu2020selfsupervised,drover20183d,kudo2018unsupervised,chenunsupervised19}, and surface correspondences estimation \cite{bhatnagar2020loopreg}. Self-supervised approaches for 3d pose estimation have primarily been developed for single-person scenarios. Given 2d poses, estimated by utilizing advances in the 2d pose estimation methods \cite{cao2017realtime,kreiss2019pifpaf,newell2017associative,cheng2020bottom,xiao2018simple,fang2017rmpe,chen2018cascaded,sun2019deep,mcnally2020evopose2d,mao2021fcpose}, these approaches use the supervisory signals generated from multi-view geometry \cite{kocabas2019self}, video constraints \cite{kundu2020selfsupervised}, or adversarial learning \cite{drover20183d,kudo2018unsupervised,chenunsupervised19}. 

Our work proposes a learning-based approach to model the 3d poses as bottleneck representations and recover geometrically constrained and spatially accurate 2d joints and heatmap representations in an end-to-end differentiable manner.

%% file: sec/3_method.tex
\section{Methodology}

\begin{figure*}[t!]
    \centering
	\includegraphics[width=1.0\linewidth]{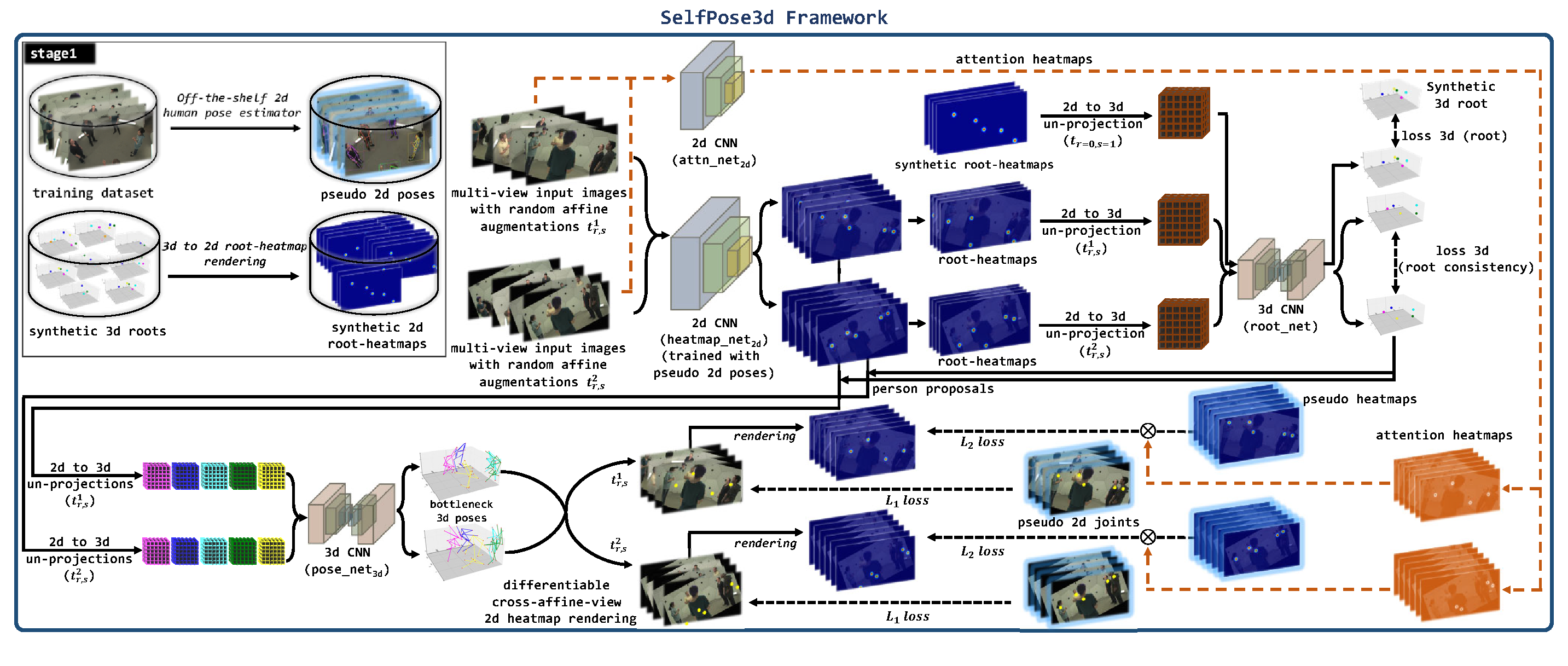}
	\caption{Illustrating our self-supervised SelfPose3d approaches for multi-view multi-person 3d pose estimation. Instead of using ground-truth 3d poses for learning, we propose self-supervised learning objectives to localize 3d roots (mid-hip location of the person) and estimate their 3d poses. We utilize a synthetic 3d roots dataset, two different affine transformations on the multi-view input images ($t_{r,s}^1, t_{r,s}^2$ parametrized by rotation $r$ and scale $s$), a differentiable cross-affine-view 2d joints and heatmaps rendering from the bottleneck 3d poses, and an adaptive supervision attention mechanism to automatically learn the 3d poses in world-space.}
	\label{fig:method}
\end{figure*}

\subsection{Problem overview} Given a training dataset of multi-view images $\mathcal{D} = \left\{ \mathbf{x}| \mathbf{y^*} \right\}$ where $ \mathbf{x} \in \mathcal{R}^{C \times 3 \times H \times W}$ is a multi-view image set from $C$ cameras with height $H$ and width $W$, and $\mathbf{y^*} \in \mathcal{R}^{C \times P \times J \times 2}$ represents the 2d pseudo poses for $P$ persons with $J$ joints, the goal is to learn a deep learning model that estimates the 3d poses $\mathcal{Y} \in \mathcal{R}^{P \times J \times 3}$ of all the $P$ persons from the multi-view input image $\mathbf{x}$. It is to be noted that $P$ can vary in each camera view due to occlusion and noisy pseudo 2d pose estimation. For simplicity in the notation, we keep the same variable P. 

Fully-supervised approaches rely on 3d ground-truth poses, while we only have 2d pseudo poses $\mathbf{y^*}$. Therefore, after obtaining 3d poses $\mathcal{Y} \in  \mathcal{R}^{P \times J \times 3}$ following traditional approach, we propose to project the poses to each view obtaining 2d poses $\mathbf{y} \in \mathcal{R}^{C \times P \times J \times 2}$, and train the model from 2d pseudo poses $\mathbf{y^*} \in \mathcal{R}^{C \times P \times J \times 2}$. 

In the following, we present our self-supervised approach based on fully-supervised VoxelPose~\cite{tu2020voxelpose}. We first generate pseudo 2d poses, and then propose \emph{self-supervised 3d root localization}, \emph{self-supervised 3d pose estimation}, and an \emph{adaptive supervision attention} to learn 3d poses in a self-supervised manner, without modifying the original VoxelPose structure.

\subsection{Generating pseudo 2d poses}
To circumvent the dependence on the ground-truth 2d poses, we generate the 2d pseudo poses on the training dataset using Mask R-CNN \cite{he2017mask} to generate person bounding boxes followed by using HRNet \cite{sun2019deep} to generate 2d poses of each detected person bounding box. The two-stage approach is chosen based on its state-of-art performance on the COCO dataset \cite{lin2014microsoft}. We first pre-train the 2d CNN backbone $\mathrm{heatmap}\_\mathrm{net}_{\mathrm{2d}}$ with pseudo 2d poses.

\subsection{Self-supervised 3d root localization}

Given 2d multi-view heatmaps from all the views and all the joints $\mathcal{H} \in \mathcal{R}^{C \times J \times \frac{H}{4} \times \frac{W}{4}}$ estimated using a 2d backbone model $\mathrm{heatmap}\_\mathrm{net}_{\mathrm{2d}}$, we use \cite{iskakov2019learnable} to construct a discretized 3d feature volume $\mathcal{F} \in \mathcal{R}^{J \times X \times Y \times Z}$ for each joint by un-projecting the 2d multi-view heatmaps to 3d space:
\begin{equation}
	\label{eq:proj}
	\mathcal{P}_{\mathrm{unproj}}(\mathrm{cam},\mathrm{center},t_{r,s})\colon \mathcal{H}  \longrightarrow \mathcal{F},
\end{equation}

To localize persons' root (mid-hip) joint in 3d space without using 3d ground truth, we hypothesize that 2d multi-view heatmaps of the root location $\mathcal{H}_{\mathrm{root}} \in \mathcal{R}^{C \times \frac{H}{4} \times \frac{W}{4}}$ are sufficient for 3d root localization (see \cref{sec:root_heatmap} for verification). Then we generate the 3d feature volume for the root location $\mathcal{F}_{\mathrm{root}} \in \mathcal{R}^{X \times Y \times Z}$ using $\mathcal{H}_{\mathrm{root}}$ using \cref{eq:proj}, which has the same dimensions as predicted root-volumes $\mathcal{G}$. This allows us to establish a one-to-one relationship between 3d root-volumes $\mathcal{G}$ and 2d multi-view root-heatmaps $\mathcal{H}_{\mathrm{root}}$. We generate a synthetic root dataset $\mathcal{D}_{\mathrm{root}} = \left\{{\mathcal{G}_{i}^{\mathrm{syn}*}}| \mathcal{H}_{\mathrm{root}\_i}^{\mathrm{syn}} \right\}_{i=1}^{N}$ where $\mathcal{G}_{i}^{\mathrm{syn}*} \in \mathcal{R}^{X \times Y \times Z}$ contains the root-volumes of randomly placed 3d points, and $\mathcal{H}_{\mathrm{root}\_i}^{\mathrm{syn}}$ is the corresponding 2d multi-view heatmaps generated by projecting the random 3d points to each view using camera parameters $\mathrm{cam}$. After unprojecting $\mathcal{H}_{\mathrm{root}}^{\mathrm{syn}}$ to $\mathcal{F}_{\mathrm{root}}^{\mathrm{syn}}$, and passing it through $\mathrm{root}\_net$ obtaining $\mathcal{G}^{\mathrm{syn}}$, we compute the $L_2$ loss error as the synthetic root loss $l_{\mathrm{root}\_\mathrm{syn}}$: 

\begin{equation}
 l_{\mathrm{root}\_\mathrm{syn}} = \mathcal{L}_2(\mathcal{G}^{\mathrm{syn}}, \mathcal{G}^{\mathrm{syn}*})
 \label{eq:root_syn}
\end{equation}

To further regularize $\mathrm{root}\_\mathrm{net}$ on the real-world 2d multi-view root-heatmaps, we propose the root consistency loss. Given a multi-view training image set $x^0$, we apply two affine transformations ($t_{r,s}^1, t_{r,s}^2$) with random rotation and scaling ($r,s$) to generate two affine transformed multi-view images ($x^1$, $x^2$). We pass $x^0$, $x^1$ and $x^2$ through $\mathrm{heatmap}\_\mathrm{net}_{\mathrm{2d}}$, construct the root feature volumes using \cref{eq:proj} with corresponding affine transformation parameters, and finally obtain $\mathcal{G}^0$, $\mathcal{G}^1$ and $\mathcal{G}^2$ through $\mathrm{root}\_\mathrm{net}$. Since $\mathcal{G}^1$ and $\mathcal{G}^2$ are invariant to the applied affine transformations $t_{r,s}^1$ and $t_{r,s}^2$, we use $\mathcal{G}^0$ as the baseline to compute the $L_2$ loss error
between $\mathcal{G}^0$, $\mathcal{G}^1$ and $\mathcal{G}^2$ as the root consistency loss $l_{\mathrm{root}\_C}$: 

\begin{equation}
 l_{\mathrm{root}\_C} = \mathcal{L}_2(\mathcal{G}^{0}, \mathcal{G}^{1}) + \mathcal{L}_2(\mathcal{G}^{0}, \mathcal{G}^{2})
 \label{eq:root_consistency}
\end{equation}

We train $\mathrm{root}\_\mathrm{net}$ by minimizing \cref{eq:root_syn} and \cref{eq:root_consistency}. We generate the person proposals $\left\{{\mathrm{root}_i}\right\}_{i=1}^{K}$ by applying non-maximum suppression (NMS) and thresholding on $\mathcal{G}^2$ ($\mathcal{G}^1$ would also work).

\subsection{Self-supervised 3d pose estimation}
Given pseudo 2d poses $\mathbf{y}^*_{\mathrm{2d}}$, person proposals $\left\{{\mathrm{root}_i}\right\}_{i=1}^{P}$, and 2d multi-view heatmaps $\mathcal{H}^{1}, \mathcal{H}^{2}$ predicted using $\mathrm{heatmap}\_\mathrm{net}_{\mathrm{2d}}$, we describe our self-supervised 3d pose estimation approach.

The person proposals $\left\{{\mathrm{root}_i}\right\}_{i=1}^{P}$ are used to generate the 3d feature volumes \emph{i.e}  $\left\{{\mathcal{F}_i^1}\right\}_{i=1}^{P} = \left\{{\mathcal{P}_{\mathrm{unproj}}(\mathrm{cam},r_i,t_{r,s}^1)}\right\}_{i=1}^{P}$  and $\left\{{\mathcal{F}_i^2}\right\}_{i=1}^{P} = \left\{{\mathcal{P}_{\mathrm{unproj}}(\mathrm{cam},r_i,t_{r,s}^2)}\right\}_{i=1}^{P}$ corresponding to the person feature volumes for each affine augmented multi-view input image $x^1$ and $x^2$, respectively. $\left\{{\mathcal{F}_i^1}\right\}_{i=1}^{P}$ and $\left\{{\mathcal{F}_i^2}\right\}_{i=1}^{P}$ are passed through $\mathrm{pose}\_\mathrm{net}_{\mathrm{3d}}$ and soft-argmax \cite{chapelle2010gradient} to estimate the 3d poses $\mathcal{Y}^1, \mathcal{Y}^2 \in  \mathcal{R}^{P \times J \times 3}$. These 3d poses serve as a bottleneck representation.

Given the camera parameters $\mathrm{cam}$ and the affine transformation parameters $t_{r,s}^1, t_{r,s}^2$, we project the bottleneck 3d poses in cross-affine-view \ie $\mathcal{Y}^1$ are projected to $x^2$ image space using $t_{r,s}^2$ to generate multi-view 2d poses $\mathbf{y}^2 \in \mathcal{R}^{C \times P \times J \times 2}$, and $\mathcal{Y}^2$ are projected to the $x^1$ image space using $t_{r,s}^1$ to generate multi-view 2d poses $\mathbf{y}^1 \in \mathcal{R}^{C \times P \times J \times 2}$.

We propose to render $\mathbf{y}^1$ and $\mathbf{y}^2$ 2d poses in the 2d heatmap representation. The heatmap representation encodes the per-pixel likelihood of a body joint and has been a vital component to enable the state-of-the-art 2d pose estimation approaches \cite{sun2019deep}. Generating heatmap representation has essentially been a pre-processing step where state-of-the-art approaches quantize the 2d joints before generating the heatmap \cite{sun2019deep}. However, this quantization step is non-differentiable and could cut the backward gradient flow. Zhang \etal \cite{zhang2020distribution} show that encoding floating point 2d joints into heatmap representation in their pre-processing step leads to improved performance. We propose to use the same differentiable approach in an online way to render the projected 2d joints into the heatmap representation.

We render $\mathbf{y}^1$ and $\mathbf{y}^2$ into heatmap representation to generate 2d multi-view heatmaps $\mathcal{H}^{1}$, and $\mathcal{H}^{2}$, respectively. We apply the affine transformations $t_{r,s}^1, t_{r,s}^2$ on the pseudo 2d poses $\mathbf{y}^*_{\mathrm{2d}}$ to generate pseudo 2d multi-view joints $\mathbf{y}^{1*}_{\mathrm{2d}}, \mathbf{y}^{2*}_{\mathrm{2d}}$ and heatmaps $\mathcal{H}^{1*}, \mathcal{H}^{2*}$. Then, we compute the $L_2$ loss between heatmaps as the pose heatmap loss $l_{\mathrm{pose}\_\mathrm{H}}$:

\begin{equation}
 l_{\mathrm{pose}\_\mathrm{H}} = \mathcal{L}_2(\mathcal{H}^{1}, \mathcal{H}^{1*}) + \mathcal{L}_2(\mathcal{H}^{2}, \mathcal{H}^{2*})
\end{equation}

After training with $l_{\mathrm{pose}\_\mathrm{H}}$ preliminarily, we propose to add the $L_1$ loss between 2d joints to further fine-tune the model. For each view, we employ the Hungarian algorithm~\cite{kuhn1955hungarian} to obtain the optimal assignment between $\mathbf{y}$ and $\mathbf{y}^*_{\mathrm{2d}}$, where the matching cost is the mean absolute error. Based on the optimal assignment, we obtain the $L_1$ loss as $l_{\mathrm{pose}\_\mathrm{J}}$. Then we use $l_{\mathrm{pose}\_\mathrm{H}}$ and $l_{\mathrm{pose}\_\mathrm{J}}$ together to train the whole network, where $\lambda$ is a manually defined weight:

\begin{equation}
 l_{\mathrm{pose}\_\mathrm{J}} = \mathcal{L}_1(\mathbf{y}^{1}, \mathbf{y}^{1*}_{\mathrm{2d}}) + \mathcal{L}_1(\mathbf{y}^{2}, \mathbf{y}^{2*}_{\mathrm{2d}}) 
\end{equation}

\begin{equation}
 l_{\mathrm{pose}\_\mathrm{3d}} = l_{\mathrm{pose}\_\mathrm{H}} + \lambda l_{\mathrm{pose}\_\mathrm{J}}
\end{equation}

As the network needs to reason about the 2d joint locations in spatial dimension, it implicitly solves the person correspondence problem. Training $\mathrm{pose}\_\mathrm{net}_{\mathrm{3d}}$ with 3d pose loss $l_{\mathrm{pose}\_\mathrm{3d}}$ performs decently, but to achieve even better results, we introduce the adaptive supervision attention. 

\subsection{Adaptive supervision attention}

\begin{figure}[t!]
    \centering
    \includegraphics[width=1.0\linewidth]{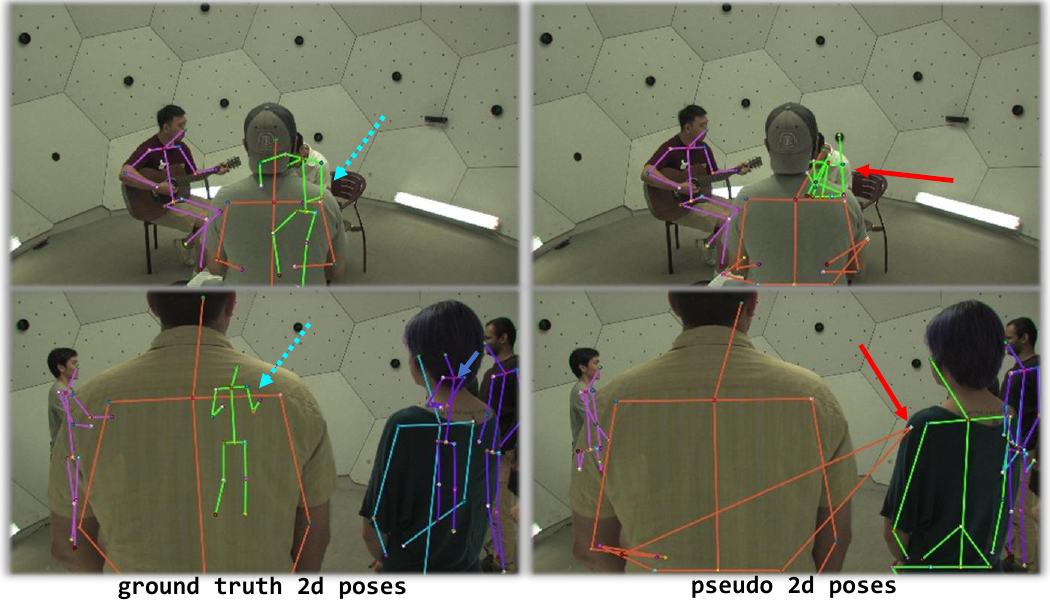}
    \caption{Comparing ground-truth 2d poses generated by projecting the ground-truth 3d poses to each multi-view image and our pseudo 2d poses generated by running HRNet human pose estimation model \cite{sun2019deep} on the training dataset. Pseudo 2d poses contain localization errors due to occlusion (see the red arrows), and ground-truth 2d poses exist for partially or even entirely occluded persons (see the blue dotted arrows).}
    \label{fig:abl_1}
\end{figure}

Traditional $L_1$ and $L_2$ losses treat each label equally, which is sub-optimal in two aspects: (1) the 2d human pose detector may generate inaccurate labels due to occlusions (see the red arrows in \Cref{fig:abl_1}); (2) the 3d-2d projection will output 2d joints in certain views even when the person is entirely occluded (see the blue dotted arrows in \Cref{fig:abl_1}). Therefore, we propose to employ attentions to adaptively guide the supervision process. 

For $L_2$ loss supervision, we use the soft attention. Specifically, we use ResNet-18 to extract the visual features of the views, followed by deconvolutional layers to obtain the attention heatmaps $\mathcal{A}$ (see $\mathrm{attn}\_\mathrm{net}_{\mathrm{2d}}$ in \Cref{fig:method}). Then we compute the element-wise product of $\mathcal{A}$ and the square error before averaging, as the new loss $l^{\mathrm{attn}}_{\mathrm{pose}\_\mathrm{H}}$: 

\begin{equation}
 l^{\mathrm{attn}}_{\mathrm{pose}\_\mathrm{H}} = \frac{1}{N} \sum_{i=1}^N \mathcal{A}_i \otimes (\mathcal{H}_i - \mathcal{H}_i^{*})^2
\end{equation}

To avoid that $\mathcal{A}$ becomes zero, we add a regularization term. We create tensors of all ones $\mathbbm{1}$ as the attention heatmap labels, and compute $L_2$ error as the attention loss $l_{attn}$. If $l_{\mathrm{attn}}$ becomes zero, $l^{\mathrm{attn}}_{\mathrm{pose}\_\mathrm{H}}$ degrades to non-attentive version. 

\begin{equation}
 l_{\mathrm{attn}} = \mathcal{L}_2(\mathcal{A}, \mathbbm{1})
\end{equation}

For $L_1$ loss supervision, we use the hard attention. For each input with $K$ views, we compute the $L_1$ loss of each view, find the view with the largest loss, and ignore it when averaging the final loss $l^{\mathrm{attn}}_{\mathrm{pose}\_\mathrm{J}}$:

\begin{equation}
 j = \mathop{\arg\max}_{i} \mathcal{L}_1(\mathbf{y}_i, \mathbf{y}^{*}_{i,\mathrm{2d}}) \quad (i=1,2,...,K)
\end{equation}

\begin{equation}
 l^{\mathrm{attn}}_{\mathrm{pose}\_\mathrm{J}} = \frac{1}{K-1} \sum_{i=1,i\neq j}^K \mathcal{L}_1(\mathbf{y}_i, \mathbf{y}^{*}_{i,\mathrm{2d}})
\end{equation}

In general, the final 3d pose loss $l^{\mathrm{attn}}_{\mathrm{pose}\_\mathrm{3d}}$ is as follows, where $\lambda$ and $\sigma$ are manually defined weights:

\begin{equation}
 l^{\mathrm{attn}}_{\mathrm{pose}\_\mathrm{3d}} = l^{\mathrm{attn}}_{\mathrm{pose}\_\mathrm{H}} + \lambda l^{\mathrm{attn}}_{\mathrm{pose}\_\mathrm{J}}  + \sigma l_{\mathrm{attn}}
 \label{eq:final_loss_pose}
\end{equation}

We train $\mathrm{pose}\_\mathrm{net}_{\mathrm{3d}}$ by minimizing $l^{\mathrm{attn}}_{\mathrm{pose}\_\mathrm{3d}}$. Our self-supervised approach is visually described in \Cref{fig:method}. 

\subsection{Implementation details}
\paragraph{\textbf{Training strategies}} For the Panoptic dataset, similar to VoxelPose \cite{tu2020voxelpose}, we first train $\mathrm{heatmap}\_\mathrm{net}_{\mathrm{2d}}$ for 20 epochs with pseudo 2d poses. We use the Adam optimizer with an initial learning rate of 1e-4, which decreases to 1e-5 and 1e-6 at the 10th and 15th epochs, respectively. Then, we train the $\mathrm{root}\_\mathrm{net}$ for $1$ epoch, followed by end-to-end joint training of the whole network for $5$ epochs using only the $L_2$ loss, with a learning rate of 1e-4. Afterwards, we add $L_1$ loss to train the whole network for another $5$ epochs with a learning rate of 5e-5. $\lambda$ and $\sigma$ in \cref{eq:final_loss_pose} are set to $0.01$ and $0.1$ respectively.

We use the random rotation between $-45^{\circ}$ to $45^{\circ}$ and random scale between $-0.35$ to $0.35$. We also apply spatial augmentations using \emph{rand-augment} \cite{cubuk2020randaugment} and \emph{rand-cutout} \cite{devries2017improved} using python image library\footnote{\url{https://github.com/jizongFox/pytorch-randaugment}}. The rand-augment consist of ``contrast-jittering'', ``auto-contrast'', ``equalize'', ``color-jittering'', ``sharpness-jittering'', and  ``brightness-jittering'', and the rand-cutout places random square boxes of sizes between 20 to 40 pixels at random locations in the image. We use the SMPL model and optimization-based body fitting approach\footnote{\url{https://github.com/JiangWenPL/multiperson/tree/master/misc/smplify-x}}\cite{loper2015smpl,bogo2016keep} to estimate body mesh parameters.

\paragraph{\textbf{Inference pipeline}} During inference, we input the multi-view RGB images, and obtain the estimated 3d poses in an end-to-end pipeline. For each view, the backbone generates corresponding 2d heatmaps for cuboid construction. Then, given constructed cuboid of the whole space, the $\mathrm{root}\_\mathrm{net}$ predicts root joint locations of all persons. Finally, the $\mathrm{pose}\_\mathrm{net}$ outputs the regressed 3d locations of each joint for every cuboid proposal of the root joints.

%% file: sec/4_experiments.tex
\section{Experiments}
\label{sec:experiments}
\subsection{Datasets and evaluation metrics}
We conduct experiments on three benchmark datasets: \emph{Panoptic} \cite{joo2015panoptic}, \emph{Campus} \cite{belagiannis20143d}, and \emph{Shelf} \cite{belagiannis20143d}.

The Panoptic dataset is a large-scale dataset captured inside a dome environment containing multiple persons performing daily social activities. We conduct extensive experiments on this dataset to evaluate and assess various components of our approach. We use the same data sequences for training and testing as VoxelPose \cite{tu2020voxelpose} except that our training set doesn't include `160906\_band3'. In other words, we are only using 9 multi-view videos for training (the `160906\_band3' video is not available due to the broken images on the source website). We use the five HD camera images (3, 6, 12, 13, 23) to train and report the performance in our experiments. We use Average Precision (AP), Recall, and Mean Per Joint Position Error (MPJPE) in millimeters (mm) as evaluation metrics (higher AP and lower MPJPE are better) \cite{tu2020voxelpose}.

The Shelf and Campus are two multi-person datasets capturing activities in the indoor and outdoor environments, respectively \cite{belagiannis20143d}. We use the same training and test split as \cite{belagiannis20143d, tu2020voxelpose}. As used in \cite{belagiannis20143d, tu2020voxelpose}, we use the Percentage of Correct Parts (PCP) as evaluation metrics for these two datasets.

\begin{figure*}[t!]
    \centering
    \includegraphics[width=0.9\linewidth]{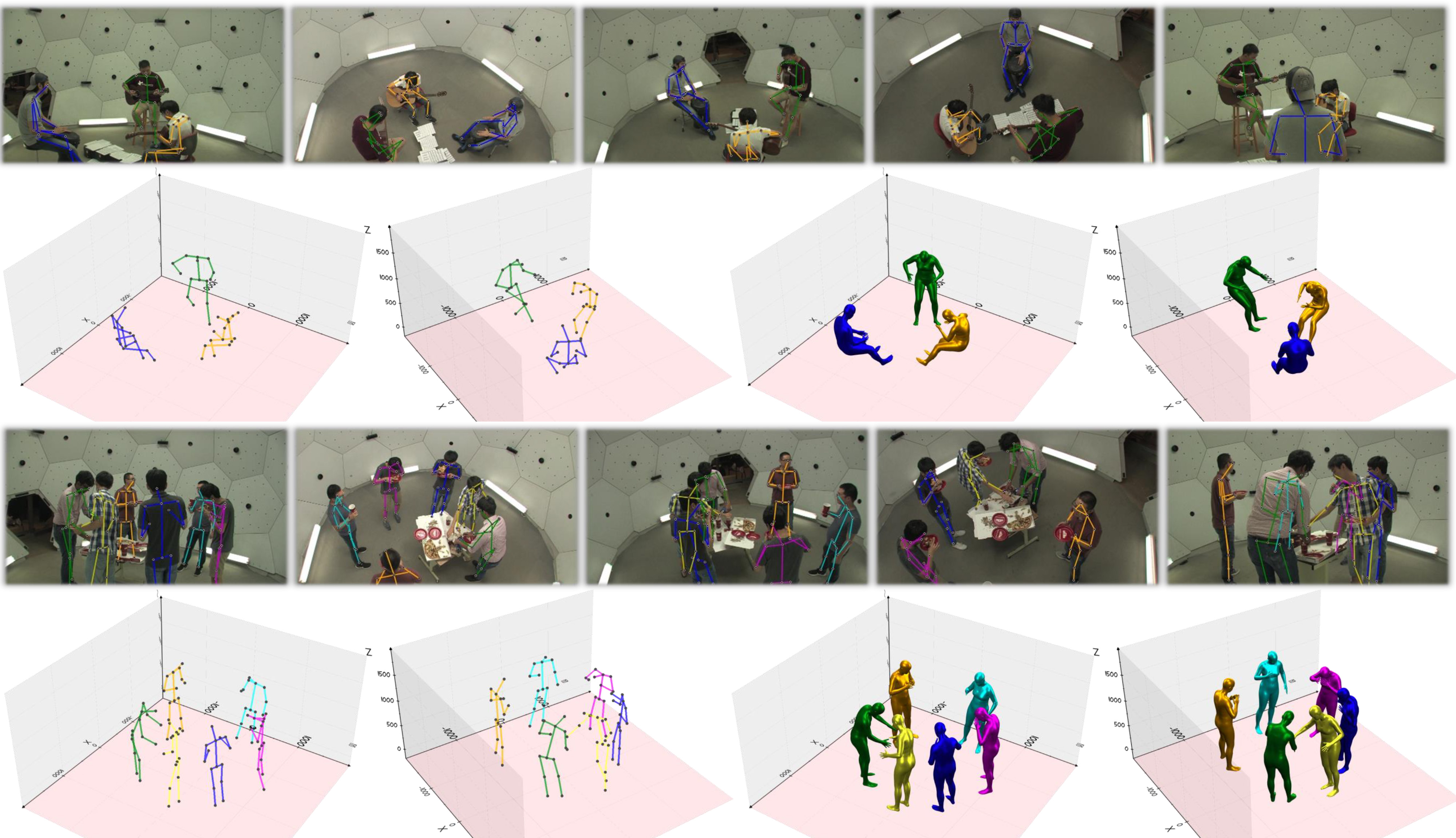}
    \caption{Qualitative results for the 3d pose estimations, 2d projections on the multi-view images, and estimated SMPL body shapes on some example images from the Panoptic dataset}. 
    \label{fig:qual_1}
\end{figure*}

\begin{table}
    \centering
    \small
    \scalebox{0.66}{
        \begin{tabular}{l|c|cccccc}
            \toprule
             & Methods                                                             & AP$_{25}$     & AP$_{50}$     & AP$_{100}$    & AP$_{150}$    & Recall$_{@500}$ & MPJPE[mm]     \\
            \midrule
            \parbox[c]{5mm}{\multirow{5}{*}{\rotatebox[origin=c]{90}{\makecell{FS}}}}
             & VoxelPose~\cite{tu2020voxelpose}              & 83.6          & 98.3          & \textbf{99.8} & \textbf{99.9} & 98.8            & 17.7          \\
             & Lin \etal~\cite{lin2021multi}                            & 92.1          & 99.0 & \textbf{99.8} & 99.8          & -               & 16.8          \\
             & MvP~\cite{zhang2021direct}                       & \textbf{92.3} & 96.6          & 97.5          & 97.7          & 98.2            & 15.8 \\
             & Wu \etal~\cite{wu2021graph} & - & - & - & - & 98.7 & 15.8 \\
             & TEMPO~\cite{choudhury2023tempo} & 89.0 & \textbf{99.1} & \textbf{99.8} & \textbf{99.9} & - & \textbf{14.7} \\
            \midrule
            \parbox[b]{5mm}{\multirow{2}{*}{\rotatebox[origin=c]{90}{\makecell{OB}}}}
             & ACTOR~\cite{pirinen2019domes}  & -             & -             & -             & -             &  -             & 168.4         \\
             &  MvPose~\cite{dong2019fast}                     & 0.0          & 2.97          &  59.93        & 81.53          & 98.23   & 84.2          \\
            \cmidrule{1-8}
            \parbox[b]{5mm}{\multirow{1}{*}{\rotatebox[origin=c]{90}{\makecell{SS}}}}
             & SelfPose3d (ours)                                                   & 55.1          & 96.4          & 98.5          & 99.0          & \textbf{99.6}   & 24.5          \\
            \bottomrule
        \end{tabular}
    }
    \caption{\small{Result on the Panoptic dataset (FS = fully-supervised, OB = optimization-based, SS = self-supervised).}} 
    \label{tab:panoptic_main}
\end{table}
\begin{table}
    \centering
    \small
    \scalebox{0.55}{
        \begin{tabular}{l|c|cccc|cccc}
            \toprule
             & \multirow{2}{*}{Methods}                             & \multicolumn{4}{c|}{Shelf} & \multicolumn{4}{c}{Campus}                                                                                                 \\ \cmidrule{3-10}
             &                                                            & Actor 1                    & Actor 2                    & Actor 3       & Average       & Actor 1       & Actor 2       & Actor 3       & Average       \\ \midrule
            \parbox[c]{5mm}{\multirow{7}{*}{\rotatebox[origin=c]{90}{\makecell{FS}}}}
             & Ershadi \textit{et al.}~\cite{ershadi2018multiple}         & 93.3                       & 75.9                       & 94.8          & 88.0          & 94.2          & 92.9          & 84.6          & 90.6          \\
             & Wu \textit{et al.}~\cite{wu2021graph}                      & 99.3                       & \textbf{96.5}              & 97.3          & \textbf{97.7} & -             & -             & -             & -             \\
             & MvP~\cite{zhang2021direct}               & 99.3                       & 95.1                       & 97.8          & 97.4          & \textbf{98.2} & \textbf{94.1} & 97.4          & 96.6          \\
             & VoxelPose~\cite{tu2020voxelpose}     & 99.3                       & 94.1                       & 97.6          & 97.0          & 97.6          & 93.8          & \textbf{98.8} & \textbf{96.7} \\
             & VoxelPose$^*$~\cite{tu2020voxelpose} & \textbf{99.5}              & 93.5                       & 97.8          & 96.9          & 93.1          & 86.5          & 93.2          & 90.9          \\
            \midrule
            \parbox[b]{5mm}{\multirow{2}{*}{\rotatebox[origin=c]{90}{\makecell{OB}}}}
             & 3DPS~\cite{belagiannis20153d}       & 75.3                       & 69.7                       & 87.6          & 77.5          & 93.5          & 75.7          & 84.4          & 84.5          \\
            & MvPose~\cite{dong2019fast}                   & 98.8                       & 94.1                       & 97.8          & 96.9          & 97.6          & 93.3          & 98.0          & 96.3          \\             
            \midrule  
            \parbox[b]{5mm}{\multirow{1}{*}{\rotatebox[origin=c]{90}{\makecell{SS}}}}
             & SelfPose3d                                       & 97.2                       & 90.3                       & \textbf{97.9} & 95.1          & 92.5          & 82.2          & 89.2          & 87.9          \\
            \bottomrule
        \end{tabular}
    }
    \caption{\small{Results (in PCP) on Shelf and Campus datasets (FS = fully-supervised, OB = optimization-based, SS = self-supervised, $^*$ = reproduced results). SelfPose3d is trained from the pseudo 3d poses from the Panoptic training set.}}
    \label{shelf_campus}
\end{table}
\paragraph{\textbf{Panoptic}} \Cref{tab:panoptic_main} shows the results on the challenging Panoptic dataset. All the fully-supervised approaches utilizing 2d and 3d ground-truth 3d poses reach nearly the same performance. SelfPose3d, without using any 3d or 2d ground-truth poses, achieves comparable results to fully-supervised approaches. Nevertheless, there still exists a gap compared to the fully-supervised VoxelPose model (96.4 \textit{v.s} 98.3 AP$_{50}$ and 24.5 \textit{v.s} 17.7 MPJPE). However, unlike VoxelPose, which relies on heat maps from all the joints to estimate 3d roots, we only use root-heatmaps to do the same. This results in the reduction of the input channel from 15 (number of keypoints) to 1 for the $\mathrm{root}\_\mathrm{net}$, making our approach computationally faster. 

We also compare our approach with optimization-based baselines from Pirinen \etal~\cite{pirinen2019domes} and Dong \etal~\cite{dong2019fast}. These non-learning-based approaches fail to capture the multi-person interaction in a complex scene from a few sparse multi-view cameras. Our learning-based self-supervised approach achieves much better performance. It is to be noted that Pirinen \etal evaluate their approach on two multi-person sequences, whereas we evaluate on four multi-person sequences.

\paragraph{\textbf{Shelf and Campus}} We compare our approach with the state-of-the-art methods on the Shelf and Campus dataset. VoxelPose uses the 3d ground-truth from the Panoptic dataset to train their approach to these datasets due to noisy and incomplete 3d ground-truth poses. For a fair comparison with VoxelPose, we use the pseudo 3d poses (by running SelfPose3d on the Panoptic training set) and train on these two datasets in a fully supervised manner. As shown in \Cref{shelf_campus}, our approach using pseudo 3d poses from the Panoptic dataset also reaches the same performance as the fully-supervised approaches.
\paragraph{\textbf{Qualitative visualizations}}
\Cref{fig:qual_1} shows 3d pose estimation results from our SelfPose3d approach on the challenging Panoptic dataset. Without using any 3d ground-truth, we can see that SelfPose3d is robust to occlusions and multiple persons while correctly identifying the person identities across all the views (see the corresponding 2d projections in \Cref{fig:qual_1}). We also show the qualitative results for the SMPL body mesh fitting \cite{loper2015smpl,bogo2016keep} on the estimated 3d poses. All these results demonstrate both the effectiveness and extendability of SelfPose3d. Please see the supplementary for more results.

\subsection{Ablation studies}
\paragraph{\textbf{Ground-truth 2d poses \textit{v.s} pseudo 2d poses}} As shown in~\Cref{tab:abl_gt2d_vs_pseudo2d}, when we use the ground-truth 2d poses in our self-supervised framework, 3d reconstruction error significantly reduces. To inspect the better performance when using the ground-truth 2d poses, we qualitatively compare the ground-truth 2d poses with the pseudo 2d poses on some example training images. Pseudo 2d poses contain localization errors due to occlusion, whereas ground-truth 2d poses exist for partially or even entirely occluded persons as shown in \Cref{fig:abl_1}. As the ground-truth 2d poses are generated by projecting the ground-truth 3d poses to each multi-view image, they serve as a suitable proxy for the 3d poses, thereby reaching a performance close to the fully-supervised approaches. However, obtaining the ground 2d poses in this way would be as challenging as acquiring the ground-truth 3d poses.

\begin{table}
  \centering
  \scalebox{0.7}{
    \begin{tabular}{c|ccc}
        \toprule
        \makecell{2d poses} & AP$_{50}$ & AP$_{100}$ & MPJPE \\
        \midrule
        ground-truth        & 98.8      & 99.6       & 19.9  \\
        pseudo              & 96.4      & 98.5       & 24.5  \\
        \bottomrule
    \end{tabular}
  }
  \caption{The ground-truth 2d poses in our self-supervised framework decrease the 3d reconstruction error and reach the performance close to the fully-supervised approaches.}
  \label{tab:abl_gt2d_vs_pseudo2d}
\end{table}

\paragraph{\textbf{Importance of cross-affine-view consistency and affine augmentations}} We also examine the effect of affine augmentations on the multi-view images and cross-affine-view consistency when generating differentiable 2d representations from the bottleneck 3d poses. As shown in \Cref{tab:cross_view_consistency}, the affine augmentations and cross-affine-view consistency significantly improve the 3d pose reconstruction as they provide necessary geometric constraints during training.

\begin{table}
  \centering
  \scalebox{0.7}{
    \begin{tabular}{c|c|ccc}
        \toprule
        \makecell{\emph{cross-affine-view} consistency} & \makecell{affine augs} & AP$_{50}$ & AP$_{100}$ & MPJPE \\
        \midrule
        \xmark     & \xmark     & 86.0 & 96.2 & 34.7 \\
        \xmark     & \checkmark & 83.3 & 97.5 & 33.3 \\
        \checkmark & \checkmark & 93.8 & 98.1 & 29.3 \\
        \bottomrule
    \end{tabular}
  }
  \caption{Affine augmentations and \emph{cross-affine-view} consistency significantly improves the 3d pose reconstruction accuracy. All three models are trained for two epochs with frozen backbone and frozen \emph{root\_net} and no attention.}
  \label{tab:cross_view_consistency}
\end{table}

\paragraph{\textbf{Analysis of $L_2$ and $L_1$ pose losses}} 
\label{l2_l1_ana}
We conduct experiments to analyze the use of $L_1$ and $L_2$ pose losses. As shown in \Cref{tab:l1_l2}, using $L_2$ and $L_1$ losses together can obtain better results than using $L_2$ loss solely. Also, using $L_1$ loss solely doesn't converge due to the label noises. 

\begin{table}
  \centering
  \scalebox{0.7}{
    \begin{tabular}{c|c|cccc}
        \toprule
        \makecell{$L_1$ loss} & \makecell{$L_2$ loss} & AP$_{25}$ & AP$_{50}$ & AP$_{100}$ & MPJPE \\
        \midrule
        \checkmark & \xmark     & - & - & - & - \\
        \xmark     & \checkmark & 43.8 & 95.8 & 98.2 & 25.7 \\
        \checkmark & \checkmark & \textbf{55.1} & \textbf{96.4} & \textbf{98.5} & \textbf{24.5} \\
        \bottomrule
    \end{tabular}
  }
  \caption{Training using $L_1$ and $L_2$ pose losses together achieves the best performance.}
  \label{tab:l1_l2}
\end{table}

\paragraph{\textbf{Importance of adaptive supervision attention}} We also examine the necessity of adaptive supervision attention. \Cref{tab:attention} shows that supervision attention for both $L_1$ and $L_2$ losses are necessary for training.

\begin{table}
  \centering
  \scalebox{0.75}{
    \begin{tabular}{c|c|cccc}
        \toprule
        \makecell{$L_1$ loss attention} & \makecell{$L_2$ loss attention} & AP$_{25}$ & AP$_{50}$ & AP$_{100}$ & MPJPE \\
        \midrule
        \xmark     & \xmark     & 32.5 & 94.1 & 97.8 & 28.5 \\
        \checkmark & \xmark     & 37.9 & 95.8 & 98.0 & 26.3 \\
        \xmark     & \checkmark & 47.4 & \textbf{96.6} & 98.2 & 25.0 \\
        \checkmark & \checkmark & \textbf{55.1} & 96.4 & \textbf{98.5} & \textbf{24.5} \\
        \bottomrule
    \end{tabular}
  }
  \caption{Training using $L_1$ and $L_2$ loss supervisions together achieves the best performance.}
  \label{tab:attention}
\end{table}

\paragraph{\textbf{Influence of different 2d human pose estimation models}} Finally, we show how pseudo 2d poses generated from different 2d human pose estimation models affect the performance. As shown in \Cref{tab:krcnn_vs_hrnet}, models that perform well on the COCO dataset \cite{lin2014microsoft} also generate better pseudo 2d poses for the Panoptic dataset, helping SelfPose3d to achieve better performance.

\begin{table}
  \centering
  \scalebox{0.65}{
    \begin{tabular}{c|cccc}
        \toprule
        \makecell{Method for 2d                                              \\pseudo pose generation} & AP$_{50}$ & AP$_{100}$ & MPJPE & \makecell{Keypoint AP on \\COCO-val\cite{lin2014microsoft}}\\
        \midrule
        Keypoint R-CNN (R-101) \cite{he2017mask} & 89.2 & 97.6 & 31.9 & 66.1 \\
        HRNet (w48 384x288) \cite{sun2019deep}   & 93.8 & 98.1 & 29.3 & 76.3 \\
        \bottomrule
    \end{tabular}
  }
  \caption{Comparing different models for generating pseudo 2d poses. Models that perform well on the COCO dataset \cite{lin2014microsoft} also generate better pseudo 2d poses for the Panoptic dataset, helping SelfPose3d to achieve better performance.} 
  \label{tab:krcnn_vs_hrnet}
\end{table}

%% file: sec/5_conclusion.tex
\section{Conclusion}
We present a self-supervised approach, called \emph{SelfPose3d}, to address the challenging problem of multi-view multi-person 3d human pose estimation. Unlike current state-of-the-art methods that use difficult-to-acquire 3d ground-truth poses to train the model, SelfPose3d requires only multi-view input images and an \emph{off-the-shelf} 2d human pose detector. We propose a novel self-supervised learning objective that aims to recover 2d joints and heatmaps under different affine transformations from the bottleneck 3d poses. We further improve the performance of our approach by integrating 
adaptive supervision attention to address the misinformation caused by the inaccurate 2d pseudo labels from the \emph{off-the-shelf} 2d human pose detector. We conduct extensive experiments on large-scale benchmark datasets, assess various components of our approach, and show that SelfPose3d reaches a performance on par with the well-established fully-supervised baselines. We visualize the 3d pose reconstruction in the complex multiple-person scenes and show that body shape meshes fitted on the estimated 3d poses look geometrically plausible under different viewpoints.


%% file: sec/X_suppl.tex
\clearpage
\setcounter{page}{1}
\maketitlesupplementary

\section{Additional Experiments}

\subsection{Effect of number of persons}

To evaluate the effect of different numbers of persons, we present the video-level results of SelfPose3d and VoxelPose on the Panoptic test set, as each video contains a different number of persons. As shown in \Cref{tab:panoptic_video}, the variance of VoxelPose's performance is larger, and there is no strong correlation between the number of persons and the models' performance. We also observe that the occlusion is still the key factor because the video ``160906\_ian5'' is of a kid playing with a woman, but he is heavily occluded due to his height, resulting in lower performance.

\begin{table}[t!]
    \centering
    \small
    \scalebox{0.58}{
        \begin{tabular}{c|c|cccc|cccc}
            \toprule
            \multirow{2}{*}{Video name} & \multirow{2}{*}{\makecell{Number of \\ persons}}  & \multicolumn{4}{c|}{SelfPose3d} & \multicolumn{4}{c}{VoxelPose} \\ \cmidrule{3-10}
             &                                                            & AP$_{25}$     & AP$_{50}$  & AP$_{100}$  & MPJPE  & AP$_{25}$     & AP$_{50}$  & AP$_{100}$  & MPJPE \\ \midrule
             160906\_ian5 & 2 & 54.1  & 86.5  & 94.1  & 25.9  & 65.7 & 85.8 & 94.4 & 24.0 \\
             160422\_haggling1 & 3 & 56.0 & 95.3 & 98.0 & \textbf{23.8} & 86.2 & 98.0 & 99.5 & 17.2 \\
             160906\_band4 & 3 & \textbf{58.6} & \textbf{98.9} & 99.0 & 24.7 & \textbf{98.1} & \textbf{99.6} & 99.8 & \textbf{15.4} \\
             160906\_pizza1 & 6 & 48.6  & 97.7  & \textbf{99.7} & 24.7 & 71.3 & 98.5 & \textbf{99.9} & 20.7 \\
            \midrule
             All videos & 2-6 & 55.1 & 96.4 & 98.5 & 24.5 & 81.8 & 98.0 & 99.4 & 18.3 \\
            \bottomrule
        \end{tabular}
    }
    \caption{Video-level test results on the Panoptic dataset.}
    \label{tab:panoptic_video}
\end{table}

\subsection{Cross-scene generalization}

To test the cross-scene generalization ability of SelfPose3d, we compare it with fully-supervised VoxelPose~\cite{tu2020voxelpose} and MvP~\cite{zhang2021direct} in two directions.

\textbf{From Panoptic to Campus/Shelf.} In this part, SelfPose3d and VoxelPose are trained on the Panoptic dataset with 5 views, and then tested on the Campus and Shelf dataset without fine-tuning. For MvP, we use the provided best models. As shown in \Cref{tab:P2SC_generalization}, SelfPose3d performs better than VoxelPose and MvP, showing better cross-scene generalization from a large dataset to a smaller dataset. The significant gap on the Campus dataset also shows that SelfPose3d is more robust to the number of camera views. 

\textbf{From Campus/Shelf to Panoptic.} For SelfPose3d, we show the self-supervised learning result on the Panoptic dataset as it requires no 3D ground-truth labels. For VoxelPose and MvP, since they cannot be trained with Campus and Shelf datasets because of smaller dataset size and the noisy 3D ground-truth labels, we follow the original papers' training strategy, i.e., for VoxelPose, training using the synthetic Campus/Shelf dataset by randomly placing 3d poses of the Panoptic dataset in the Campus/Shelf 3D space; for MvP, using the provided MvP model, first trained on the Panoptic dataset and then fine-tuned on Shelf dataset. We test the above VoxelPose and MvP model, trained on the Campus/Shelf datasets, back on the Panoptic test set. As shown in \Cref{tab:SC2P_generalization}, VoxelPose and MvP fail to detect any 3d pose, although they have used 3D ground-truth labels from the Panoptic dataset in the first place. In other words, they are severely overfitted on the camera poses of the training set. The experiment shows the ability of SelfPose3d to address large-scale unseen datasets.

\begin{table}
    \centering
    \small
    \scalebox{0.63}{
        \begin{tabular}{c|cccc|cccc}
            \toprule
             \multirow{2}{*}{Methods}  & \multicolumn{4}{c|}{Shelf (5 camera views)} & \multicolumn{4}{c}{Campus (3 camera views)}  \\ \cmidrule{2-9}
              & Actor 1  & Actor 2 & Actor 3 & Average & Actor 1 & Actor 2 & Actor 3   & Average \\ \midrule
            VoxelPose & 99.5 & 93.5  & 97.8 & 96.9  & 93.1 & 86.5 & 93.2  & 90.9 \\
            VoxelPose$^*$ & 94.6  & 91.4 & 97.5 & 94.5  & 0.0 & 0.3 & 0.0 & 0.1  \\
            \midrule
            MvP & 99.3 & 95.1 & 97.8 & 97.4 &98.2 & 94.1 & 97.4 & 96.6 \\
            MvP$^*$ & 3.51  & 4.32 & 15.9 & 7.91  & 0.41 & 0.05 & 0.43 & 0.30 \\
            \midrule
             SelfPose3d  & 93.7 & 94.3  & 97.7 & 95.2 & 78.2 & 8.0 & 40.9 & 42.3 \\
            \bottomrule
        \end{tabular}
    }
    \caption{Results (in PCP) on Shelf and Campus test set without fine-tuning. (1) SelfPose3d is trained on the Panoptic dataset without using GT labels. (2) VoxelPose and MvP are with fine-tuning, and VoxelPose$^*$ and MvP$^*$ are without fine-tuning.}
    \label{tab:P2SC_generalization}
\end{table}

\begin{table}
  \centering
  \scalebox{0.6}{
    \begin{tabular}{c|cccc}
        \toprule
        Methods & AP$_{25}$ & AP$_{50}$ & AP$_{100}$ & MPJPE \\
        \midrule
        VoxelPose (Campus) & 0.0   & 0.0   & 0.0  & inf   \\
        VoxelPose (Shelf)  & 0.0   & 0.0   & 0.0  & 350.6  \\
        MvP (Shelf)        & 0.0   & 0.0   & 0.0  & 395.3 \\\midrule
        SelfPose3d         & 55.1  & 96.4  & 98.5 & 24.5   \\
        \bottomrule
    \end{tabular}
  }
  \caption{Results on the Panoptic test set. (1) VoxelPose is trained on synthetic Campus/Shelf dataset. (2) MvP is firstly trained on the Panoptic dataset and then fine-tuned on Shelf dataset. (3) SelfPose3d is trained on the Panoptic dataset in a self-supervised way.}
  \label{tab:SC2P_generalization}
\end{table}

\subsection{Ablation study on adding $L_1$ joint loss}

\begin{figure*}
    \centering
    \includegraphics[width=0.85\linewidth]{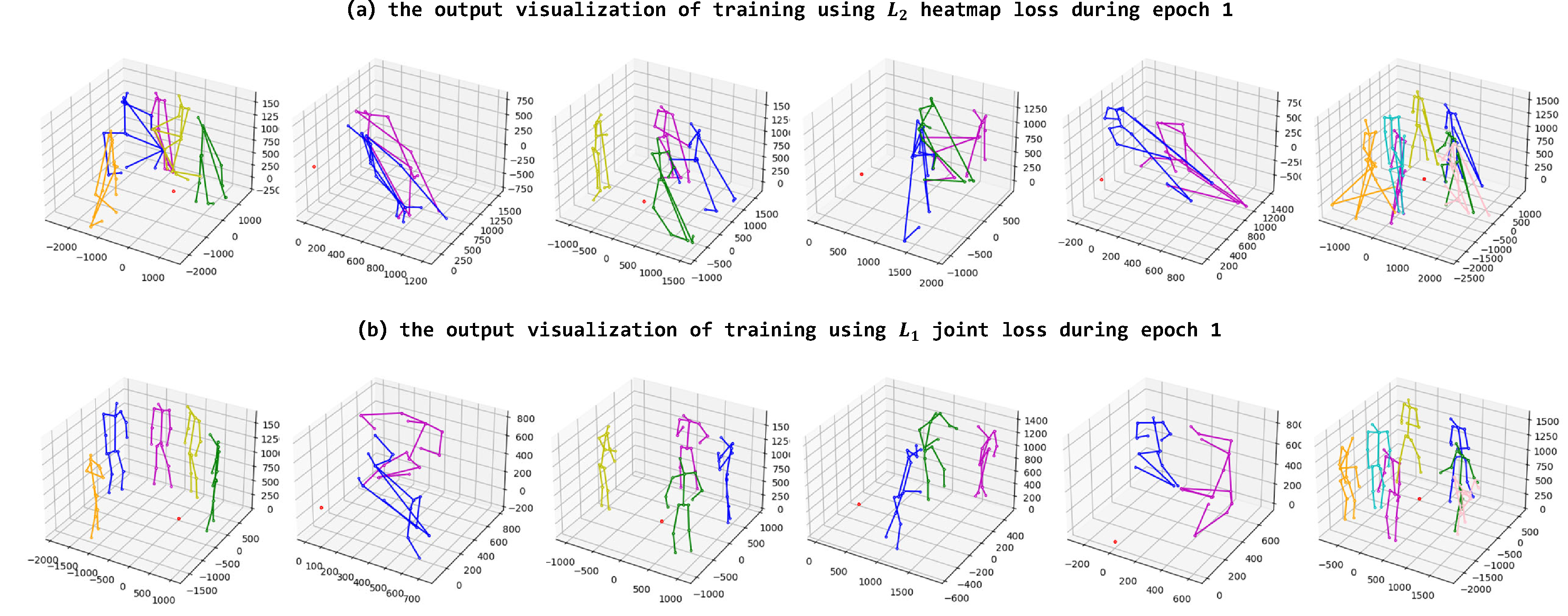}
    \caption{Comparing the visualization of the output 3d poses during epoch 1, using $L_2$ heatmap loss and $L_1$ joint loss respectively.}
    \label{fig:vis_l2_l1}
\end{figure*}

As mentioned in \cref{l2_l1_ana}, it is more likely to diverge when training the model using $L_1$ joint loss solely. However, based on the visualization of the output 3d poses in the training process (see \Cref{fig:vis_l2_l1}), we find that $L_1$ loss can help the model generate a human-shape pose much faster than $L_2$ loss in the early training stage. It is reasonable because $L_1$ loss provides a direct supervision on joint coordinates while $L_2$ loss doesn't. Thus we assume that $L_1$ loss is helpful for more precise prediction, and conduct an ablation study on merging it with $L_2$ loss in \Cref{tab:lambda}. Based on the results, we set $\lambda$ in \cref{eq:final_loss_pose} to 0.01. 

\begin{table}
  \centering
  \scalebox{0.6}{
    \begin{tabular}{c|cccc}
        \toprule
        \makecell{$\lambda$} & AP$_{25}$ & AP$_{50}$ & AP$_{100}$ & MPJPE \\
        \midrule
        0.001       & 28.4   & 93.5      & 97.4      & 28.7     \\
        0.01        & \textbf{33.6} & \textbf{95.1} & \textbf{97.7} & \textbf{27.7} \\
        0.1         & 21.8   & 79.2      & 92.4      & 34.0     \\
        1.0         & 2.71   & 44.0      & 85.3      & 48.3     \\
        \bottomrule
    \end{tabular}
  }
  \caption{Ablation study on $\lambda$ in \cref{eq:final_loss_pose}, where we train each model for 5 epochs without adding $L_1$ loss attention.}
  \label{tab:lambda}
\end{table}

\subsection{Ablation study on $L_2$ loss attention}

There are two aspects affecting the supervision attention for $L_2$ loss: the weight $\sigma$ of $l_{\mathrm{attn}}$ in \cref{eq:final_loss_pose} and the backbone. We first use ResNet-18 as the backbone, and conduct experiments about $\sigma$ in \Cref{tab:attn_sigma}. When we set $\sigma$ to 0.01, the model doesn't converge because the output of $\mathrm{attn}\_\mathrm{net}_{\mathrm{2d}}$ is almost zero. Therefore, we set $\sigma$ in \cref{eq:final_loss_pose} to 0.1. 

Afterwards, we try to deepen the architecture of $\mathrm{attn}\_\mathrm{net}_{\mathrm{2d}}$ backbone, and examine whether $\mathrm{attn}\_\mathrm{net}_{\mathrm{2d}}$ and $\mathrm{heatmap}\_\mathrm{net}_{\mathrm{2d}}$ can share weights. \Cref{tab:attn_net} shows that ResNet-18 is sufficient, and sharing weights degrades the performance. 

\begin{table}
  \centering
  \scalebox{0.6}{
    \begin{tabular}{c|cccc}
        \toprule
        \makecell{$\sigma$} & AP$_{25}$ & AP$_{50}$ & AP$_{100}$ & MPJPE \\
        \midrule
        0.01        & -      & -      & -      & -     \\
        0.1         &\textbf{36.6} &\textbf{95.1} & \textbf{97.9} &\textbf{26.6}   \\
        1.0         & 32.5   & 94.3   & 97.7   & 27.6     \\
        \bottomrule
    \end{tabular}
  }
  \caption{Ablation study on $\sigma$ in \cref{eq:final_loss_pose}, where we train each model for 5 epochs using $L_2$ loss solely with ResNet-18 based $attn\_net_{2d}$.}
  \label{tab:attn_sigma}
\end{table}

\begin{table}
  \centering
  \scalebox{0.6}{
    \begin{tabular}{c|cccc}
        \toprule
        \makecell{Backbone} & AP$_{25}$ & AP$_{50}$ & AP$_{100}$ & MPJPE \\
        \midrule
        ResNet-18        & 36.6          & 95.1          & \textbf{97.9} & \textbf{26.6} \\
        ResNet-34        & \textbf{37.2} & \textbf{95.2} & 97.7          & 26.9     \\
        ResNet-50        & 26.9          & 91.6          & 97.4          & 29.9     \\
        ResNet-50$^*$    & 24.5          & 91.9          & 97.4          & 30.2     \\
        \bottomrule
    \end{tabular}
  }
  \caption{Ablation study on the backbone network of $attn\_net_{2d}$, where we train each model for 5 epochs using $L_2$ loss solely with $\sigma$=0.1. $^*$ means shared backbone with $heatmap\_net_{2d}$.}
  \label{tab:attn_net}
\end{table}

\subsection{Robustness of SelfPose3d}

In order to test the robustness of our methods, we train SelfPose3d using fewer camera views of the Panoptic dataset. As shown in \Cref{tab:views}, the performance of SelfPose3d steadily reduces when we decrease the number of camera views to 3. 

\begin{table}
  \centering
  \scalebox{0.8}{
    \begin{tabular}{c|ccccc}
        \toprule
        \makecell{Methods} & Views & AP$_{25}$ & AP$_{50}$ & AP$_{100}$ & MPJPE \\
        \midrule
        VoxelPose~\cite{tu2020voxelpose} & 5 & 83.6 & 98.3 & 99.8 & 17.7 \\
        VoxelPose~\cite{tu2020voxelpose} & 3 & 58.9 & 93.9 & 98.4 & 24.3 \\
        \midrule
        SelfPose3d (ours)    & 5 & 55.1 & 96.4 & 98.5 & 24.5 \\
        SelfPose3d (ours)    & 4 & 31.1 & 89.6 & 96.7 & 30.2 \\
        SelfPose3d (ours)    & 3 & 10.4 & 66.1 & 90.4 & 43.5 \\
        \bottomrule
    \end{tabular}
  }
  \caption{Results on the Panoptic dataset with different numbers of camera views.}
  \label{tab:views}
\end{table}

\subsection{Root localization with only root-heatmaps}
\label{sec:root_heatmap}

We use the similar architecture compared to VoxelPose for our SelfPose3d approach. The only architectural change in the SelfPose3d w.r.t VoxelPose is using only the root-heatmaps as input to the $\mathrm{root}\_\mathrm{net}$ for root localization. This architectural change has enabled us to learn the $\mathrm{root}\_\mathrm{net}$ parameters from synthetic 3d roots. \Cref{tab:abl1_root_heatmap} shows the results for root localization using only the root-heatmaps \textit{v.s} all the heatmaps for VoxelPose and SelfPose3d. We observed a minor decrease in the performance for both the approaches, confirming our hypothesis that using only 2d root-heatmaps is sufficient for 3d root localization.

\begin{table}
  \centering
  \scalebox{0.8}{
    \begin{tabular}{c|c|ccc}
        \toprule
        Method     & \makecell{ \emph{root\_net}                                   input} & AP$_{50}^{root}$ & AP$_{100}^{root}$ & MPJPE$^{root}$ \\
        \midrule
        VoxelPose  & all heatmaps                                                         & 41.0             & 99.0              & 49.3           \\
        VoxelPose  & root-heatmaps                                                        & 34.0             & 99.0              & 50.0           \\
        \midrule
        SelfPose3d & root-heatmaps                                                        & 35.2             & 92.3              & 54.9          \\
        \bottomrule
    \end{tabular}
  }
  \caption{The rationale for using root-heatmaps as input to the \emph{root\_net} for 3d roots localization. Training VoxelPose model with only root-heatmaps obtains nearly the same performance. SelfPose3d trained using synthetic root-heatmaps with root consistency loss also reaches comparable performance. Here AP$_{50}^{root}$, AP$_{100}^{root}$, and MPJPE$^{root}$ are calculated only for the root joint.}
  \label{tab:abl1_root_heatmap}
\end{table}

\begin{figure*}[t!]
    \centering
    \includegraphics[width=0.9\linewidth]{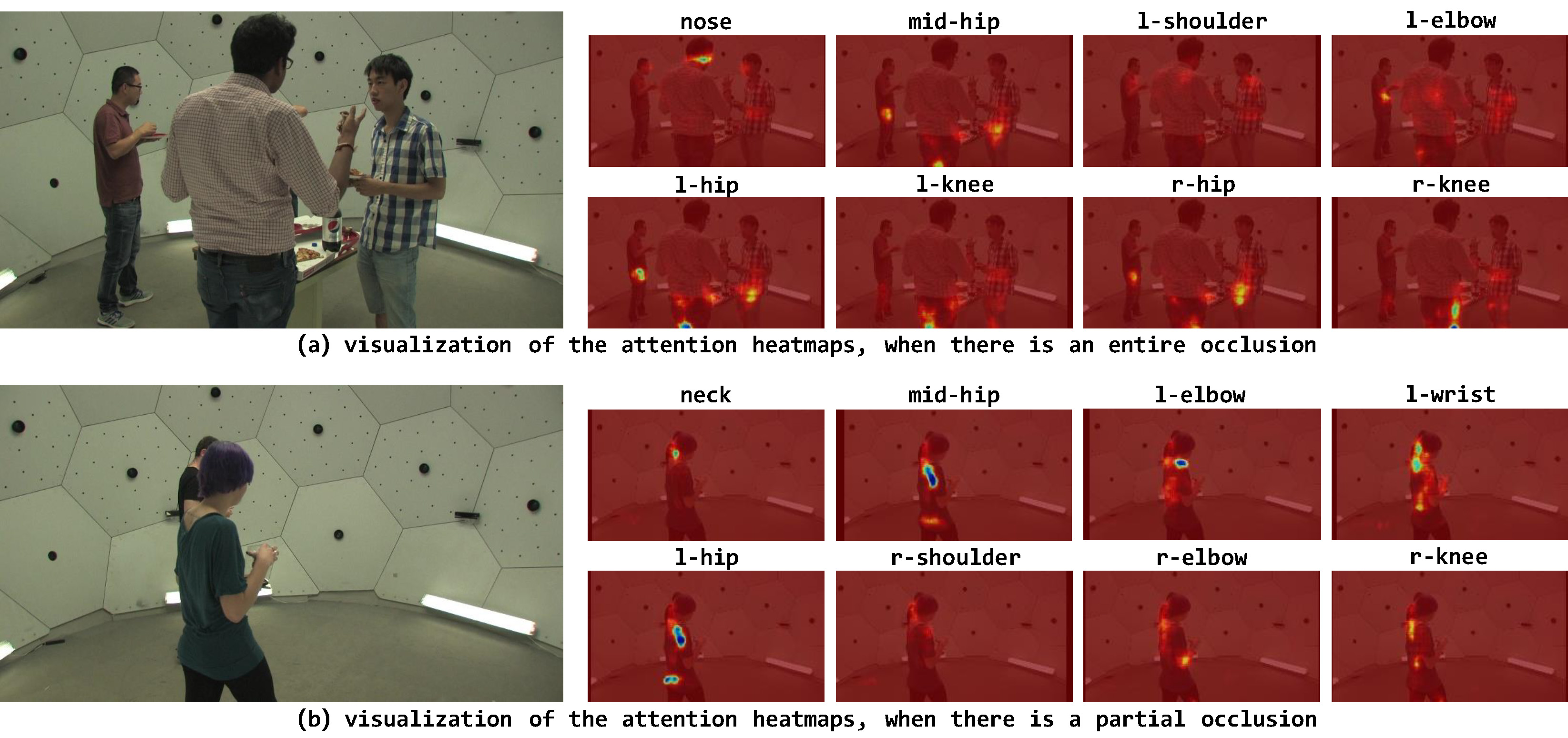}
    \caption{Visualization of the attention heatmaps. (a) The man in front of the suited man is entirely occluded, and we barely see the attention heatmaps focus on him. (b) The man is partially occluded, as we can see his head, shoulder and arm. The attention heatmaps are trying to infer the occluded part (\eg mid-hip).}
    \label{fig:attn_vis}
\end{figure*}

\subsection{Attention heatmap visualization}

To have a clearer view of the role that the $\mathrm{attn}\_\mathrm{net}_{\mathrm{2d}}$ plays in SelfPose3d, we visualize the attention heatmaps of certain views in \Cref{fig:attn_vis}. When there's an entire occlusion, $\mathrm{attn}\_\mathrm{net}_{\mathrm{2d}}$ tends to ignore the occluded person. When there's a partial occlusion, $\mathrm{attn}\_\mathrm{net}_{\mathrm{2d}}$ tends to infer the occluded part. The visualization explains the better performance when adding adaptive supervision attention. 

\subsection{Confidence threshold for pseudo labels}

To investigate whether we need to filter out the pseudo labels with low confidence scores, we generate two sets of labels: the ones with no confidence threshold are called the soft labels, and the ones with a 0.7 confidence threshold on the joints are called the hard labels. We train our model with each label set under the same experiment setting, and the results are shown in \Cref{tab:confidence}. Our main takeaways are: (1) the model trained with hard labels performs slightly better at the end (especially on the AP$_{25}$ index); (2) however, the model is more likely to collapse when we train it with hard labels. Therefore, we propose to train the model with soft labels at the beginning, and then fine-tune it with hard labels in the last 2 epochs. \Cref{tab:confidence} shows that the proposed strategy can obtain the best result, with a stable training process. 

\begin{table}
  \centering
  \scalebox{0.6}{
    \begin{tabular}{c|c|cccc}
        \toprule
        Method  & \makecell{Pseudo label \\ category} & AP$_{25}$ & AP$_{50}$ & AP$_{100}$ & MPJPE \\
        \midrule
        {\multirow{3}{*}{\makecell{SelfPose3d}}}  & soft  & 51.6 & \textbf{96.7}   & \textbf{98.6}   & 24.8   \\
         & hard  & 54.2 & 96.4  & \textbf{98.6}    & 24.6  \\
         & soft \& hard  & \textbf{55.1}  & 96.4 & 98.5  & \textbf{24.5} \\
        \bottomrule
    \end{tabular}
  }
  \caption{Comparing the models trained with (1) soft pseudo labels solely, (2) hard pseudo labels solely, and (3) two sets of labels, respectively. 
  For the soft \& hard training, we only use the hard labels in the last 2 epochs.}
  \label{tab:confidence}
\end{table}

\subsection{Failure cases}

\begin{figure}
    \centering
    \includegraphics[width=1.0\linewidth]{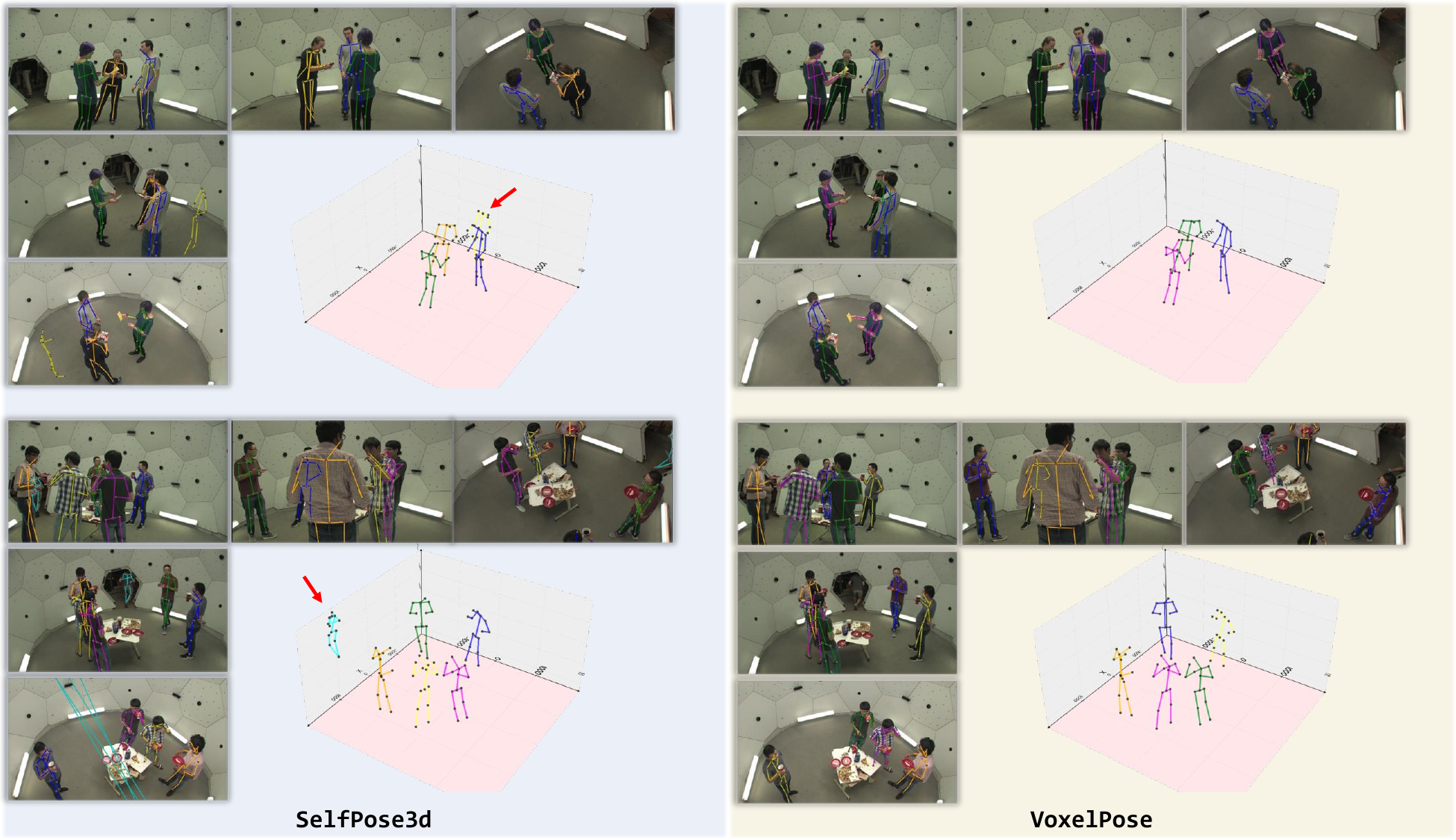}
    \caption{Failure cases from our approach compared to fully-supervised VoxelPose. The top row shows the two 3d poses for a single person, and the bottom row shows the 3d pose for a person outside the dome. Best viewed in color.}
    \label{fig:qual_compare}
\end{figure}

\Cref{fig:qual_compare} shows some failure cases from our approach compared to the fully-supervised VoxelPose. Top row of \Cref{fig:qual_compare} shows two 3d poses for a single person. Pseudo 2d poses used in our approach contain the poses of the people outside the dome, whereas the ground truth 2d and 3d poses are curated to remove the persons outside the dome. Therefore, our approach tries to infer 3d poses for the persons outside the dome (see bottom row of \Cref{fig:qual_compare}).

